\pdfoutput=1                                                          
\documentclass[a4paper, 10pt, conference]{ieeeconf}      

\IEEEoverridecommandlockouts                              
\usepackage{epsfig}           
\usepackage{mathrsfs}

\usepackage{times}
\usepackage{amsfonts}
\usepackage{amsmath}
\usepackage{epsfig}
\usepackage{subfigure}
\usepackage{ifthen}

\newcommand{\etal}{\textit{et al.}}

\newcommand{\bi}{\begin{itemize}}
\newcommand{\ei}{\end{itemize}}
\newcommand{\be}{\begin{enumerate}}
\newcommand{\ee}{\end{enumerate}}
\newcommand{\bdm}{\begin{displaymath}}
\newcommand{\edm}{\end{displaymath}}
\newcommand{\beq}{\begin{equation}}
\newcommand{\eeq}{\end{equation}}
\newcommand{\beqa}{\begin{eqnarray}}
\newcommand{\eeqa}{\end{eqnarray}}
\newcommand{\beqas}{\begin{eqnarray*}}
\newcommand{\eeqas}{\end{eqnarray*}}

\newcommand{\placeimgyy}[2]{\setlength{\epsfysize}{#1in}\epsfbox{#2}}

\title{\LARGE \bf A Mobile Robotic Personal Nightstand with Integrated Perceptual Processes}

\author{\authorblockN{Vidya N. Murali, Anthony L. Threatt, Joe Manganelli, Paul M. Yanik, Sumod K. Mohan, Akshay A. Apte,\\ Raghavendran Ramachandran, Linnea Smolentzov, Johnell Brooks, Ian D. Walker, Keith E. Green\\ }
\authorblockA{\\
Clemson University\\
Clemson, South Carolina 29634\\
Email: vmurali@clemson.edu}
}

\begin{document}
\maketitle
\thispagestyle{empty}
\pagestyle{empty}

\begin{abstract}

We present an intelligent interactive nightstand mounted on a mobile robot, to aid the elderly in their homes using physical, tactile and visual percepts.  We show the integration of three different sensing modalities for controlling the navigation of a robot mounted nightstand within the constrained environment of a general purpose living room housing a single aging individual in need of assistance and monitoring. A camera mounted on the ceiling of the room, gives a top-down view of the obstacles, the person and the nightstand. Pressure sensors mounted beneath the bed-stand of the individual provide physical perception of the person's state. A proximity IR sensor on the nightstand acts as a tactile interface along with a Wii Nunchuck (Nintendo) to control mundane operations on the nightstand. Intelligence from these three modalities are combined to enable path planning for the nightstand to approach the individual. With growing emphasis on assistive technology for the aging individuals who are increasingly electing to stay in their homes, we show how ubiquitous intelligence can be brought inside homes to help monitor and provide care to an individual. Our approach goes one step towards achieving pervasive intelligence by seamlessly integrating different sensors embedded in the fabric of the environment. 
\end{abstract}

\section{Introduction}

The most profound technologies were defined as those that disappear into the framework of the environment and become indistinguishable from the fabric of everyday life \cite{weiser1991sigmobile}. Weiser's definition then helped lay the foundation for ``pervasive computing" that deals with the distribution of intelligent sensors throughout the fabric of an environment and enabling sentient communication between them \cite{saha2008computer}. Saha \etal  also define ``pervasive computing" as follows: the mobile computing goal of ?nytime anywhere?connectivity is extended to ?ll the time everywhere?by integrating pervasiveness support technologies such as interoperability, scalability, smartness, and invisibility. It is important to note that pervasive computing is environment centric with an emphasis on ubiquitous sensing and control. 

An extensive review on pervasive computing and health care by Orwat \etal \cite{orwat2008bmc} describe that the evolving concepts of pervasive computing, ubiquitous computing and ambient intelligence are increasingly influencing health care and medicine. Because of its ubiquitous and unobtrusive analytical, diagnostic, supportive, information and documentary functions, pervasive computing is predicted to improve traditional health care. Some of its capabilities, such as remote, automated patient monitoring and diagnosis, may make pervasive computing a tool advancing the shift towards home care, and may enhance patient self-care and independent living. 

More and more members of the aging population are electing to stay in their own homes as opposed to moving into a care facility for reasons including familiarity of their homes, comfort and safety. However living alone is a daunting task for elderly people. Reflexes are not what they used to be and chronic illnesses tend to bring in risk factors. Help is needed even to perform the simplest of tasks sometimes. At night, especially it is important to have someone oversee things around the house and cater to the bedtime needs of the aging person. However this is always not possible. Psychological research has shown that older people tend to keep most of their important things cluttered onto a single night-stand that carries in it various things ranging from medicines, food, items of clothing and even bedpans and laundry \cite{green2009hri, brooks2011herdj,smoletznov2009gsa}. This venture reconceives part of an ongoing project, investigating the use of environmental sensing, inference, machine intelligence and distributed robotics to extend the independence and speed the rehabilitation of those maligned by short and long-term cognitive and physical impairments. The nightstand moves within the home in response to the individual? needs and is easy to open/close and perform other mundane tasks that may be needed by an aging person. It is the goal that this nightstand, apart from being an intelligent place to store things, should also be a device that reacts to emergency (health) situations in an appropriate manner.

\subsection{Previous Work}

The application of intelligent robotic navigation for interactive assistance in indoor environments dates back to the pioneering work by Ian Horswill \cite{horswill1993}, which describes a vision based corridor navigating robot giving tours in response to a user's demands. There are a large number of projects which could be named that show interactive design integrated with  robot navigation, some noteworthy examples being RHINO \cite{rhino} and MINERVA \cite{thrun1999minerva}. Specific applications to assisting the elderly in their homes was marked by the work of Dubowsky \etal \cite{dubowsky2000pamm}. More recently, an integrated approach to  developing a cooperative robot was described by Zender \etal \cite{zender2007aaai}. However, their work emphasizes on using natural language interface for navigation. Dialogue is critical for their navigation to be successful. However, in dealing with elderly, we have to anticipate situations where they are incapable of accurate speech, or any kind of cognitive command. In such situations, we have to rely entirely on physical and visual percepts to perform the tasks. It may also be desirable that the intelligent object work silently in a naturally reactive way without interfering with the activities of the individual. It is also important to note that it is desirable to use passive sensors as opposed to active ones like the SICK laser \cite{zender2007aaai} in an environment that may incorporate medical or sensitive equipment. Our project aims to satisfy some of these goals by using physical, tactile and visual sensing modalities described in the coming sections.

\section{Intelligent Nightstand and Testbed}

\begin{figure}
\center
\begin{tabular}{cc}
\placeimgyy{1.0}{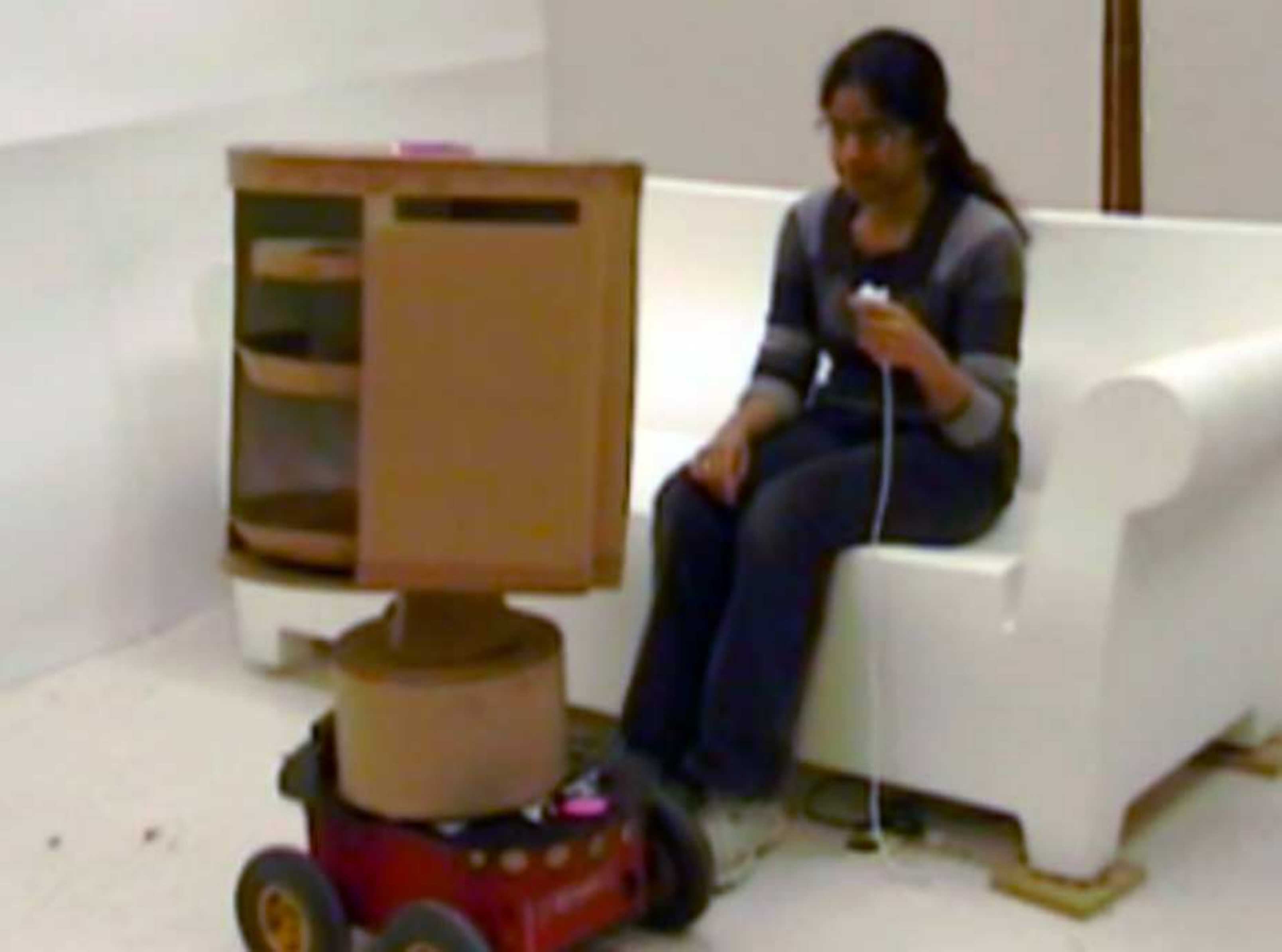}&
\placeimgyy{1.0}{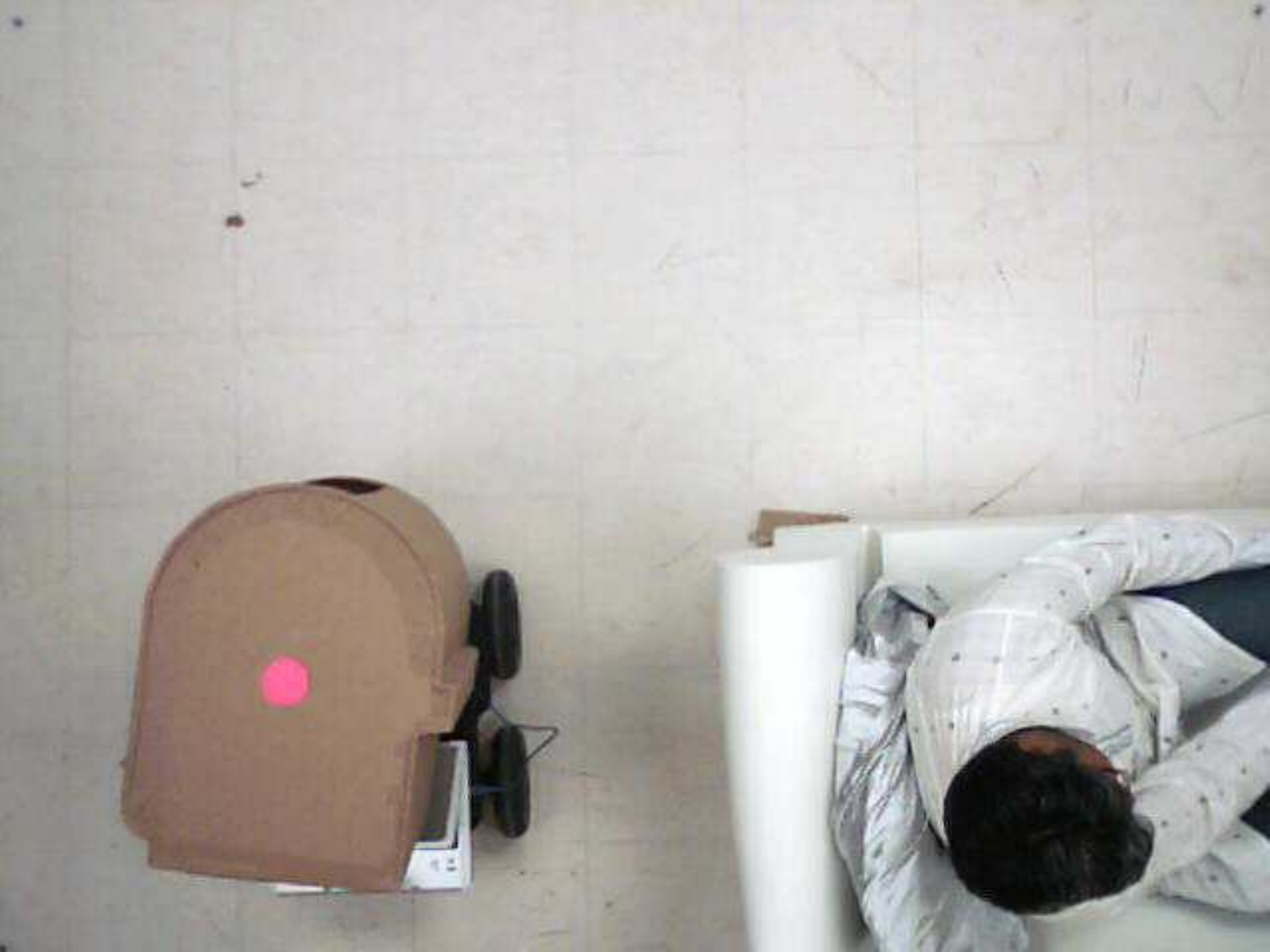}
\end{tabular}

\caption{{\sc{LEFT:}} Intelligent Nightstand mounted on the robot. {\sc RIGHT:} Top down view from the camera.}
\label{fig:testbed}
\end{figure}

We have constructed a prototype nightstand with rotating drawers and a movable screen/door using cardboard and mounted it on a ActivMedia Pioneer P3AT mobile robot platform (see Figure ~\ref{fig:testbed}). The unit is expected to move around a test lab with chairs, tables and other paraphernalia. A chair or a bed is assumed to be the rest area of the patient and has the load cells (pressure sensors) placed beneath it. A camera mounted overhead monitors the environment. A block schematic is shown (see Figure ~\ref{fig:block_schema}). For the prototype built, the camera feed and the load cells (pressure sensors) are fed into the same laptop. The nightstand inputs are fed into the laptop controlling the robot. The opening/closing of the nightstand door is controlled by an IR proximity sensor and the drawers are rotated and the nightstand is moved up/down using a Wii.	The micro controller platform used for the sensing and actuation is the ARDUINO. Arduino is an open-source electronics prototyping platform that can sense the environment by receiving input from a variety of sensors and can affect its surroundings by controlling lights, motors, and other actuators \cite{arduino}. We have used ATmega 328 and ATmega168 platforms for this venture. 
%

\section{Approach}

\subsection{Intelligence at the Room Level}

Pervasive intelligence dictates the presence of sensors in the environment. Vision is more powerful than other sensors because vision provides different kinds of information about the environment, while other sensors (such as sonars or lasers) only give us depth. For landmark detection and recognition, vision provides direct ways to do so and is easy to represent because of the close relation to the way humans understand landmarks. In addition lasers are expensive and power-hungry, and sonars cause interference. Vision based perception can now be achieved using a single off-the-shelf camera which is inexpensive and scalable. A camera is a passive sensor that can be used safely in environments sensitive to electromagnetic interference. We have deployed a single Logitech Quickcam STX webcam on the ceiling of the room under observation.  
The dominant architecture for mobile robot perception uses sensors on-board the robot, providing only a first-person perspective of the environment. But the third-person perspective from a camera mounted on the ceiling is powerful because of the inherent simplicity in dealing with data obtained from a stationary sensor \cite{hoover2000icra}. Path planning strategies are easier to implement with the god's eye view of the plan of the room, because dynamic obstacles can be learned easily and occlusions are minimized to a great extent. In our implementation this also has the advantage of monitoring the entire room continuously (pervasive).

\begin{figure}
\center
\begin{tabular}{cc}
\placeimgyy{1.0}{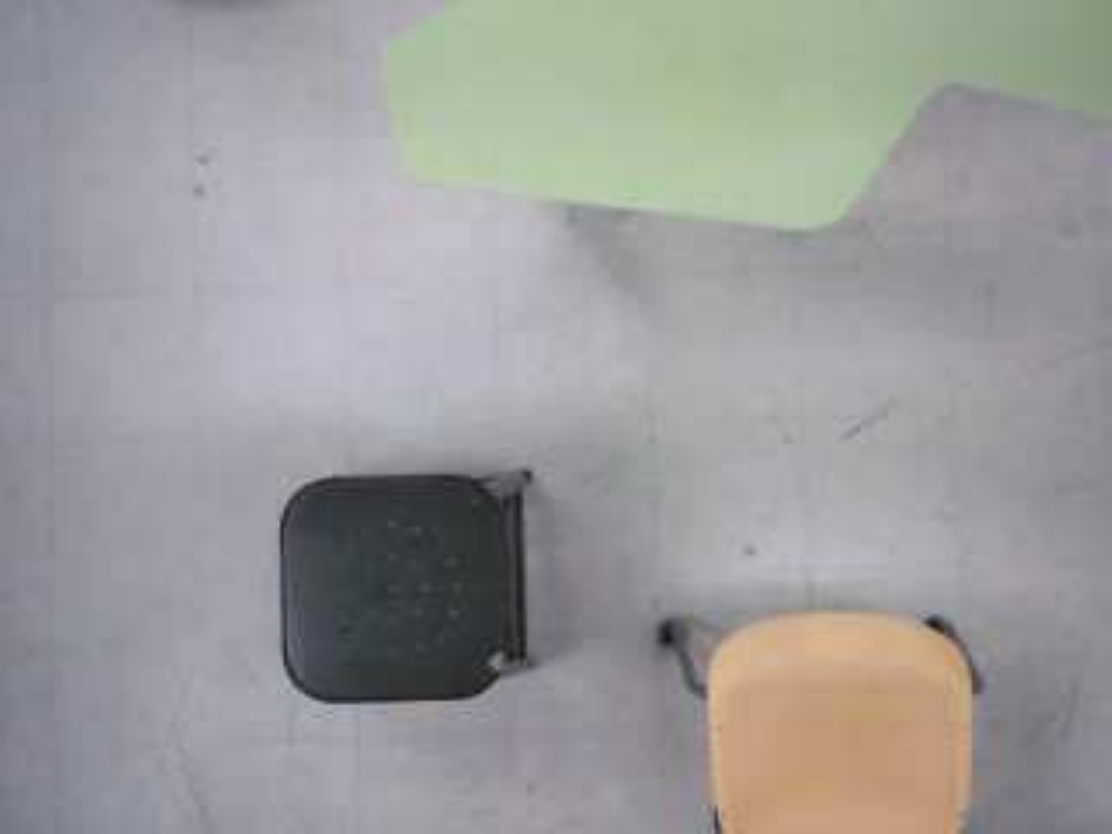}&
\placeimgyy{1.0}{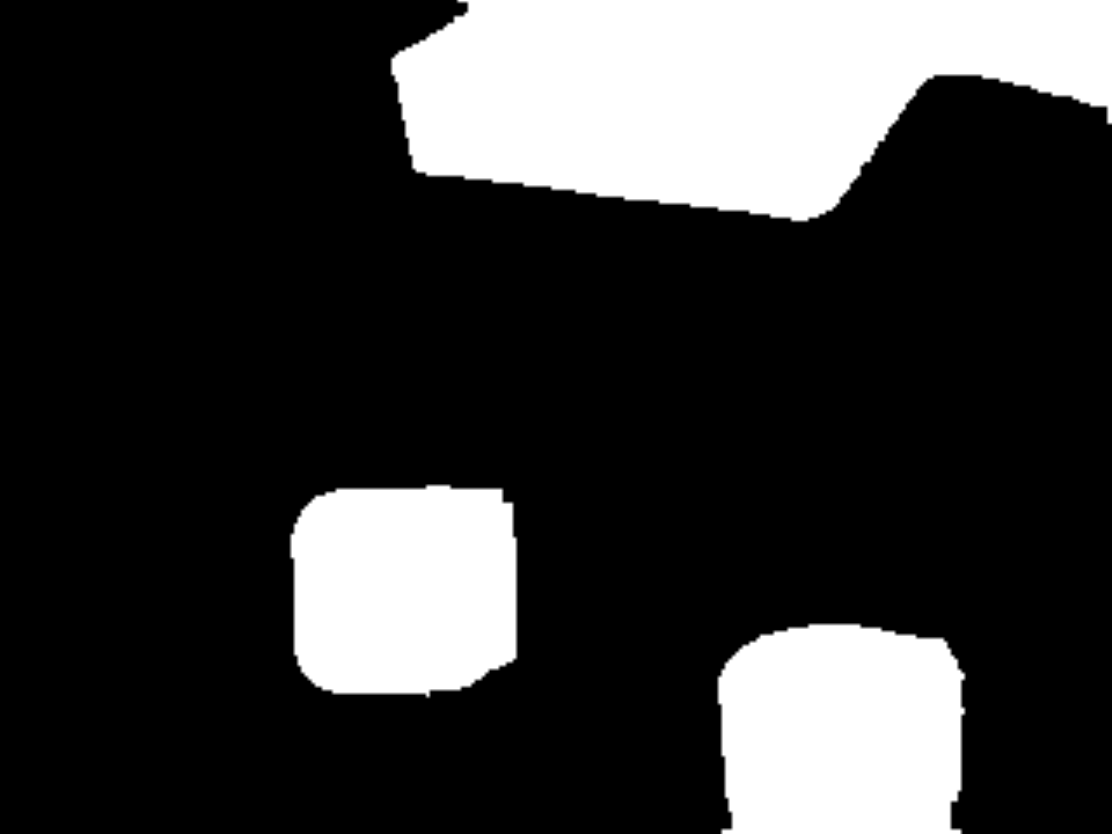}\\
Top down view & Segmented obstacles \\
\placeimgyy{1.0}{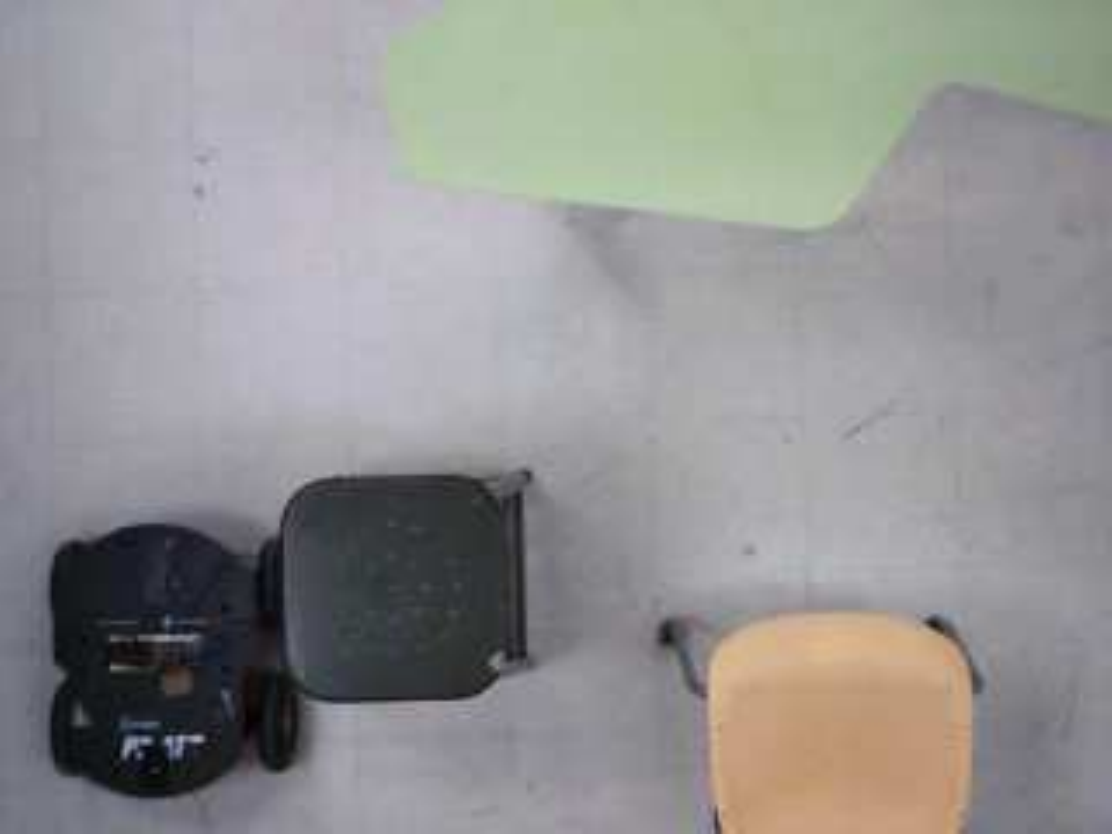}&
\placeimgyy{1.0}{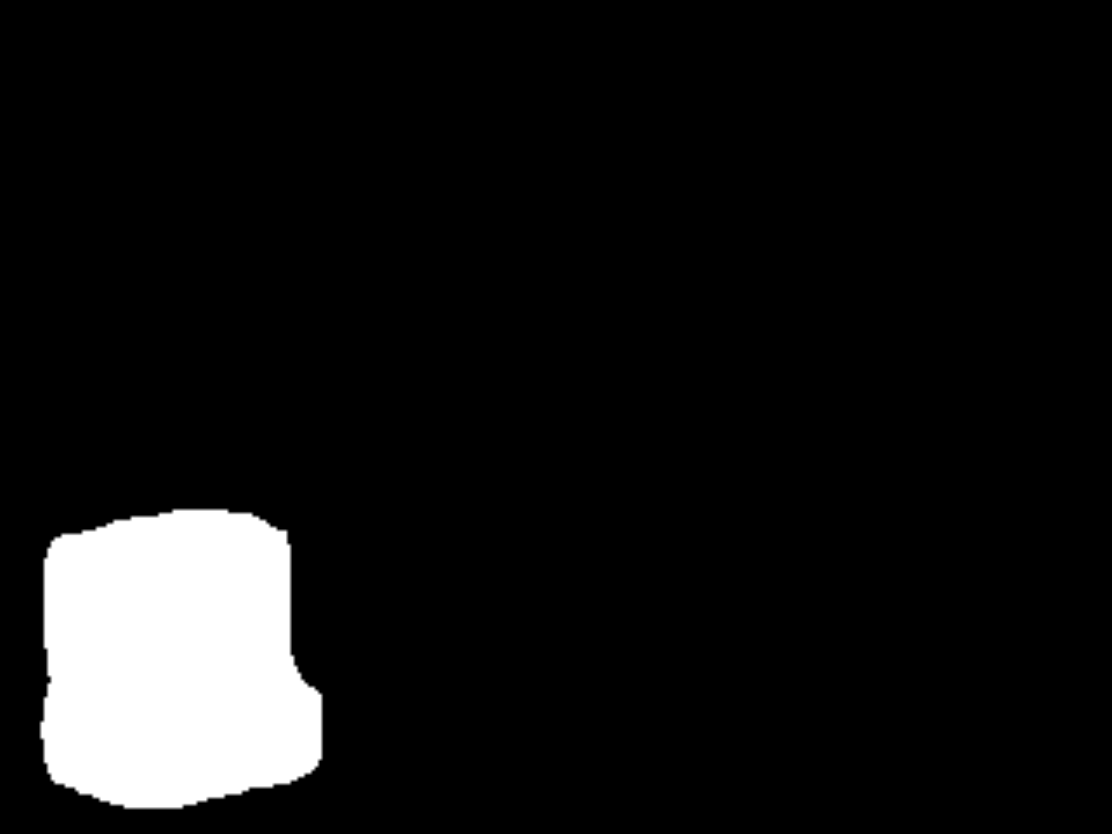}\\
Robot in view & Background subtraction \\
\placeimgyy{1.0}{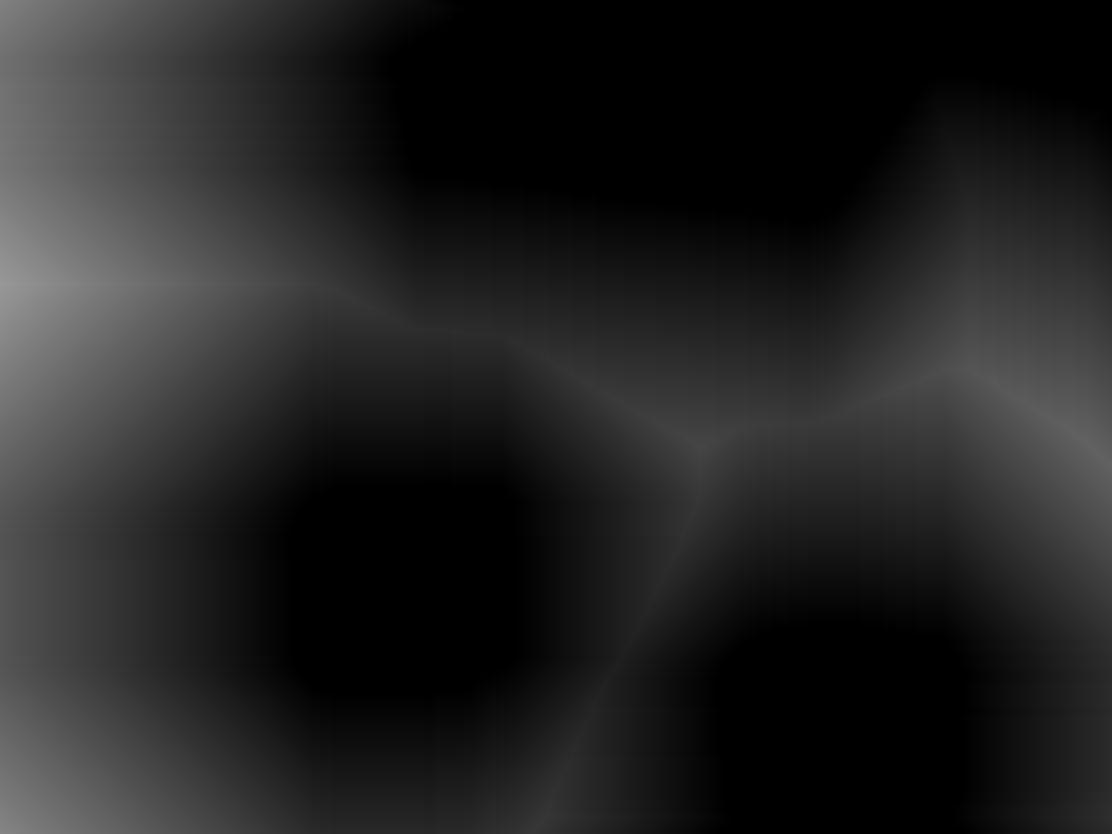}&
\placeimgyy{1.0}{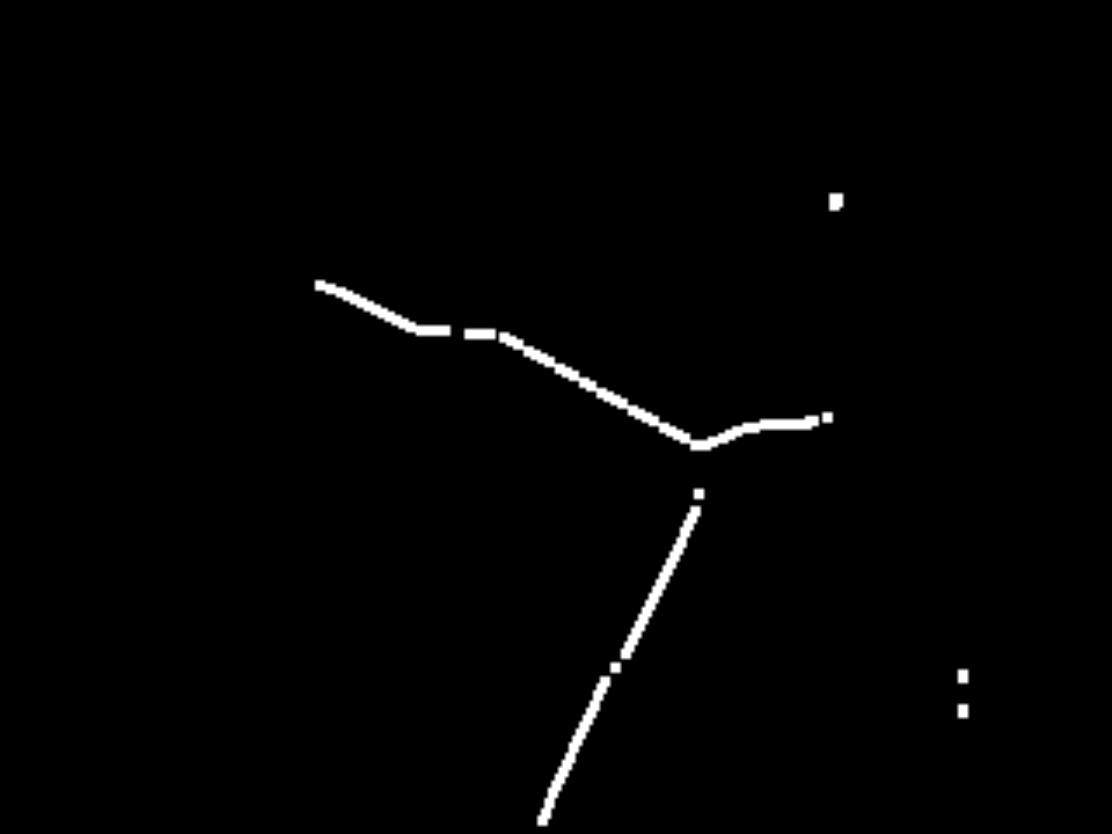}\\
Chamfer distance & Voronoi map (skeleton)

\end{tabular}
\caption{Processing of visual percepts.} 
\label{fig:vision}
\end{figure}

\subsubsection{Detection of Objects}
In the top-down image, a simple background subtraction allows us to obtain the position of the robot (on which the nightstand will be mounted). The obstacles in the room are obtained by mean-shift segmentation \cite{comaniciu2002pami}. 

\subsubsection{Path Planning for Robot Navigation}
While many complex and popular path planning algorithms are available, for the sake of simplicity we have used the Voronoi based technique. Voronoi-based maps are  roadmap methods and are preferred for indoor mapping because of their \textit{accessibility, connectivity, and departability} \cite{Choset_1997_4445} and can be constructed incrementally by the robot. In this approach, the map consists of the links which represent the obstacle-free path that can be followed by the robot. Taking the obstacle image, we first run a Chamfer distance transform \cite{Rosenfeld1968pr} on it and thin the edges to arrive at a crude Voronoi topology as shown in Figure ~\ref{fig:vision}. The links show the obstacle free paths that can be followed by the robot. 
\subsection{Intelligence Monitoring the Person}
In order to monitor physical presence and pose of the person, for this project we used pressure sensors attached to the bed stand where the individual rests. It is possible to deploy similar sensors at all resting/reclining appliances in the room. 

\subsubsection{Collection of Pose Data}

We collected readings from four pressure sensors located beneath the legs of the bed-stand with the person sitting/lying on it in various stages. The information was used to decide if the person was sitting, sleeping, one leg up and so on. The force sensor used was the FC23 compression load cell which has a maximum capacity of 500 lbs. The output of the sensors is read and processed using LabVIEW. The sensors are connected to the laptop/PC using National Instruments USB-6808. It has 8 analog and 8 digital input/outputs. It samples input voltages and converts them to numerical values, and drive output. 

\subsubsection{Analysis of Pose Data}

\begin{figure}
\center
\begin{tabular}{c}
\placeimgyy{1.75}{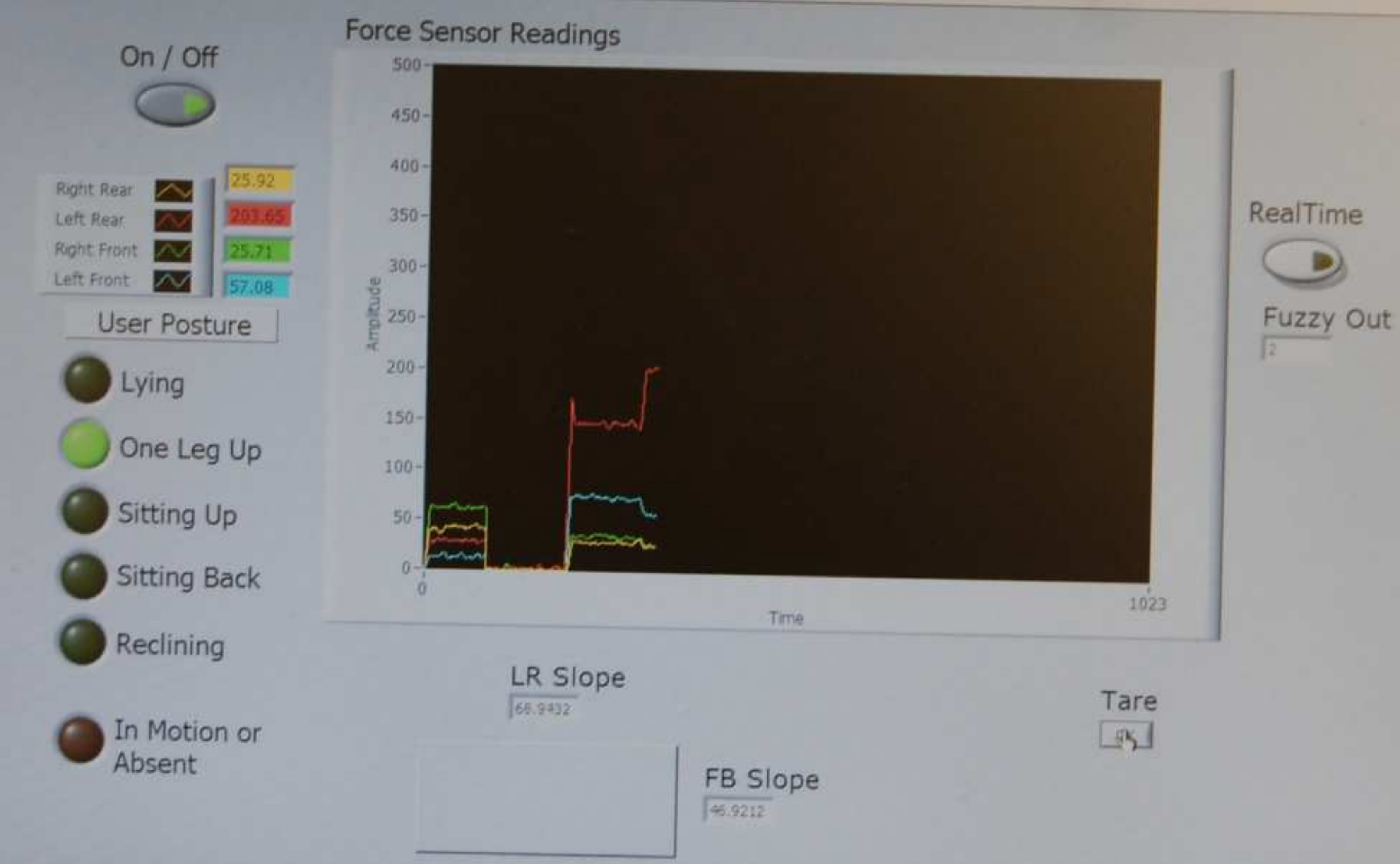}\\
\placeimgyy{1.7}{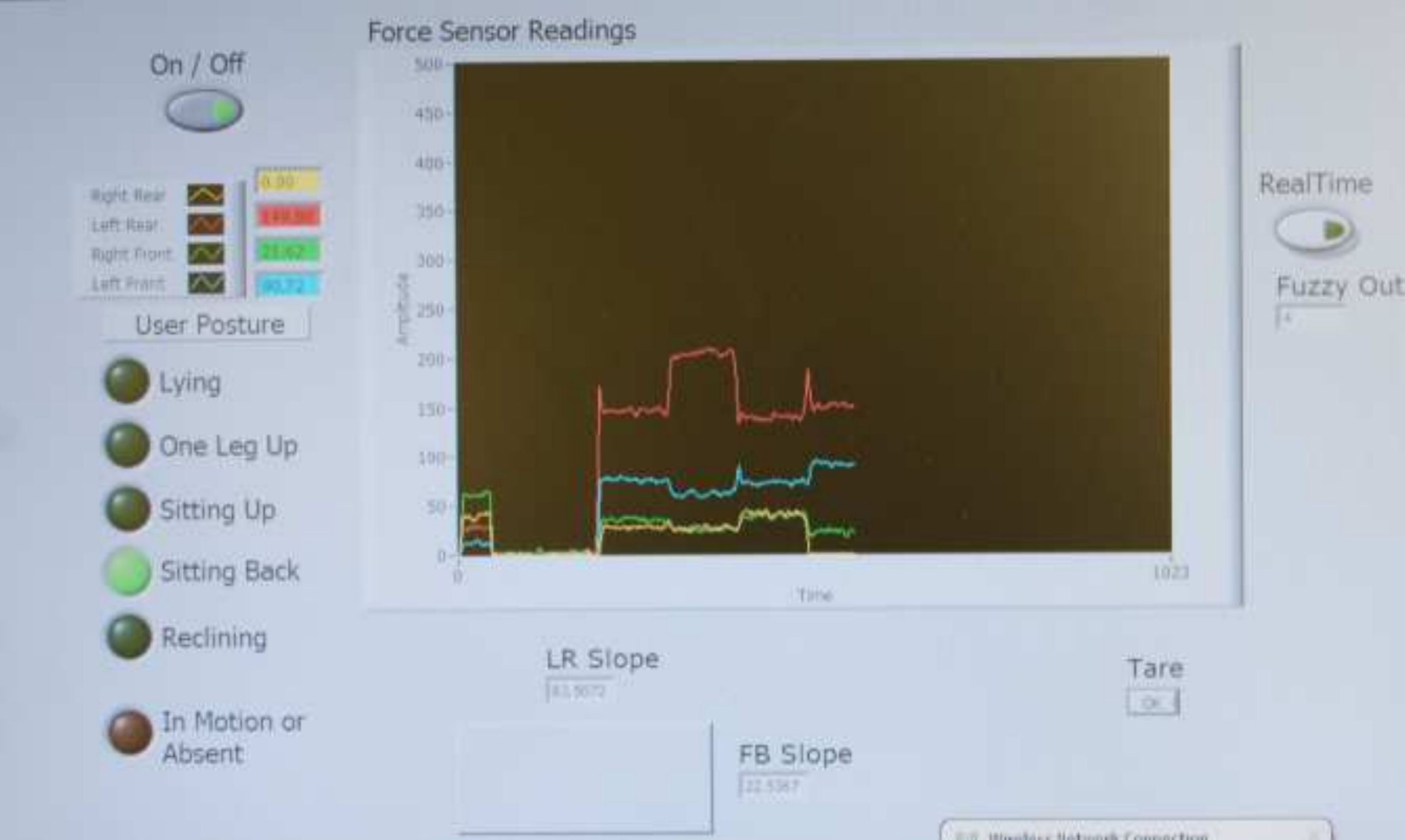}
\end{tabular}
\caption{LabVIEW output. {\sc Top:} Pose estimation for sitting with one leg up. {\sc Bottom:} Pose estimation for sitting. } 
\label{fig:poses}
\end{figure}

We used both supervised Bayesian learning and fuzzy logic to separate the data in different classes. We collected 5 sets of data for each pose as training samples. The images of the LabVIEW screen show how the data is collected in real time and also classified. See Figure ~\ref{fig:poses}. 

\subsection{Intelligence Built into the Nightstand}

\begin{figure}
\center
\begin{tabular}{cc}
\placeimgyy{1.44}{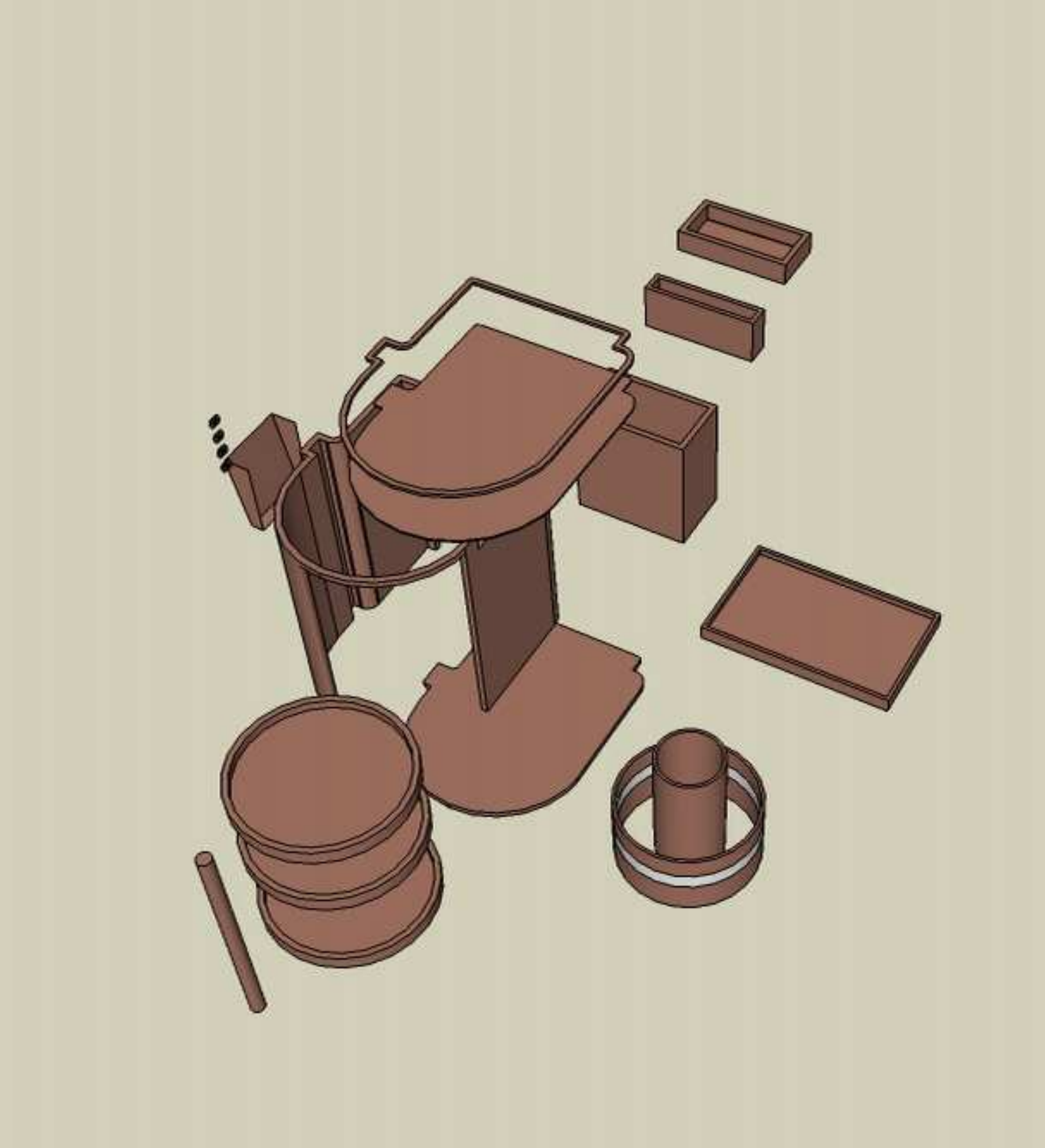}&
\placeimgyy{1.45}{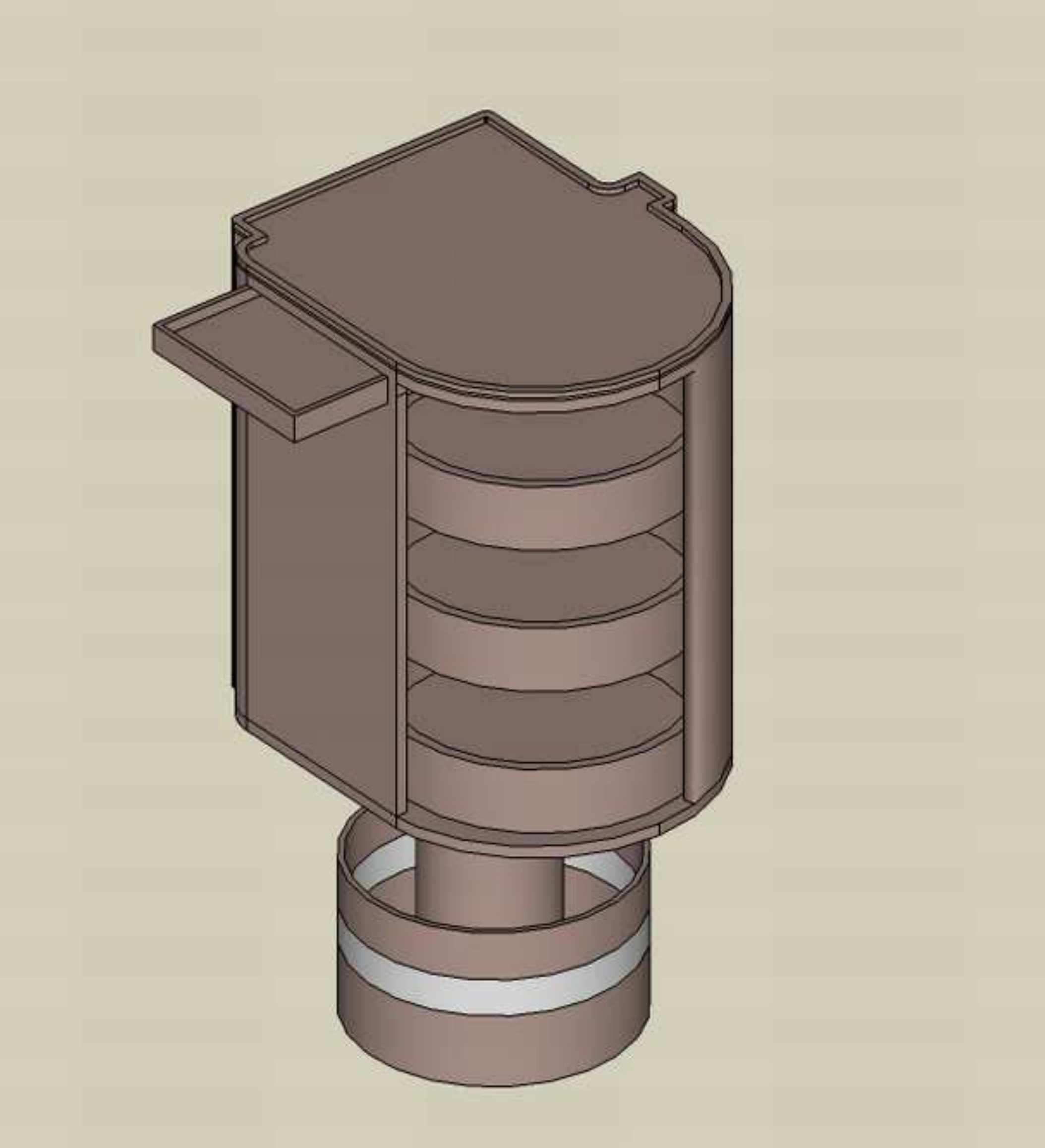}\\
\placeimgyy{1.58}{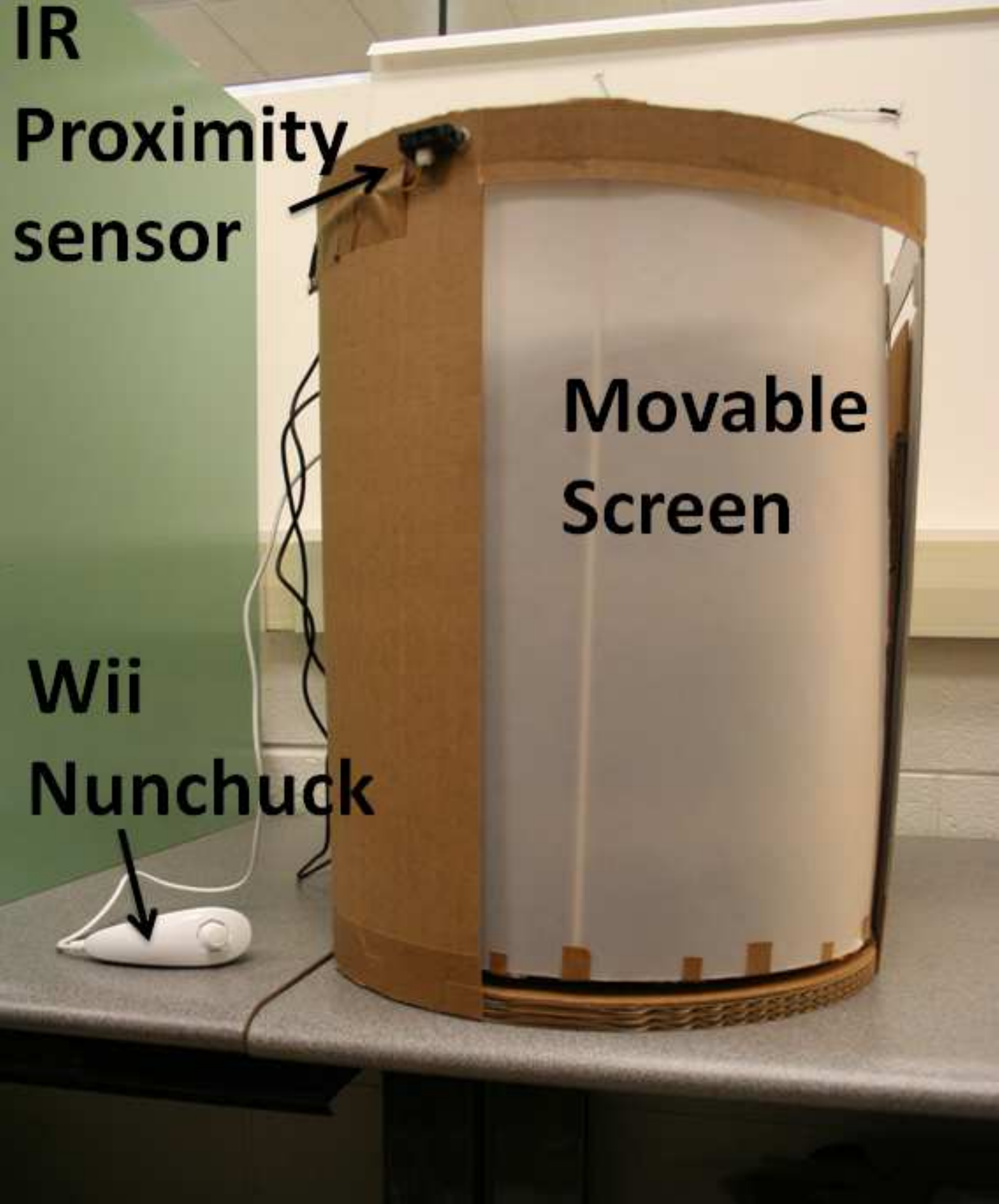}&
\placeimgyy{1.6}{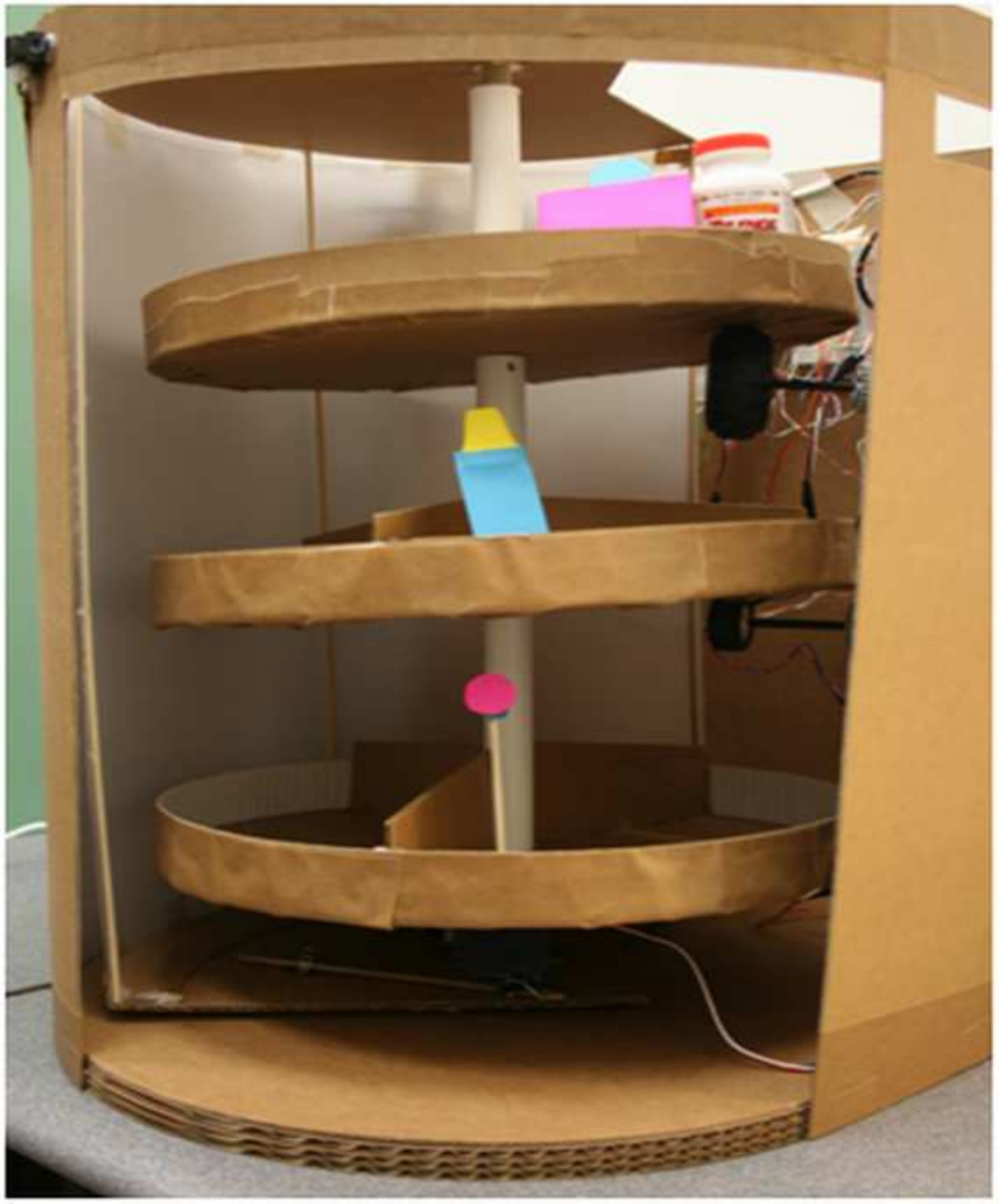}\\
\placeimgyy{0.95}{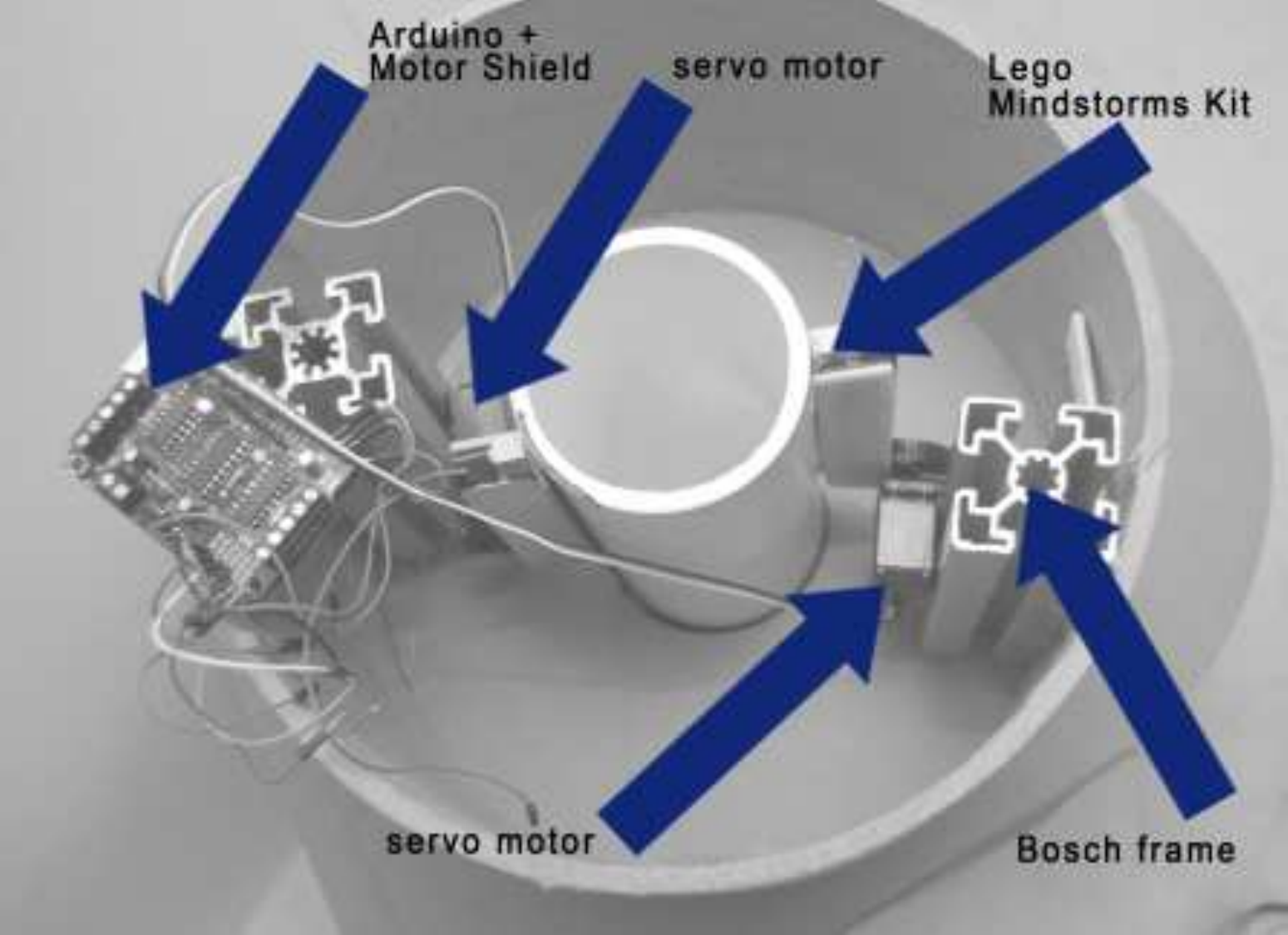}&
\placeimgyy{0.95}{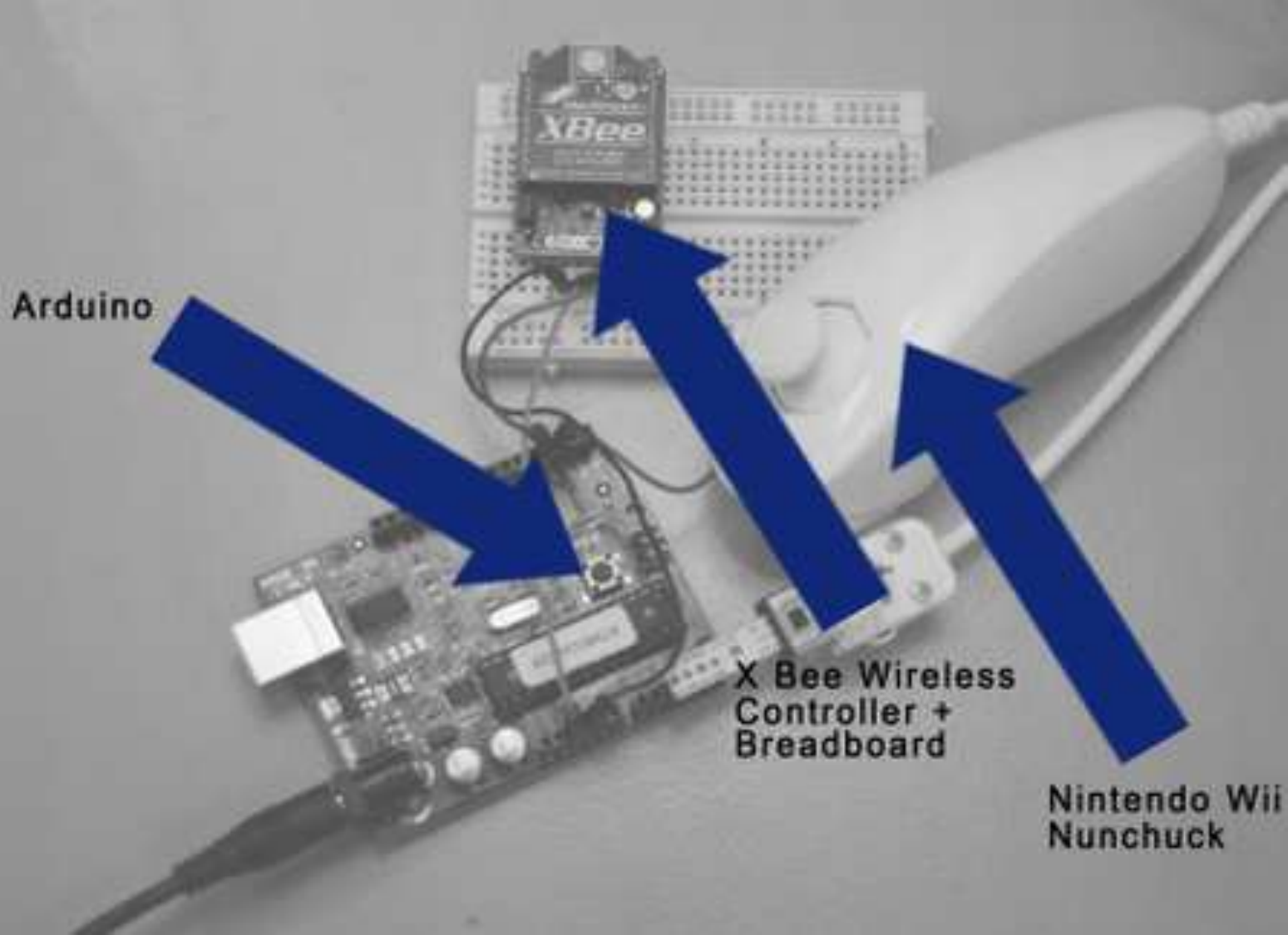}
\end{tabular}
\caption{{\sc Top:} Nightstand designs. {\sc Middle:} Nightstand with tray arrangement closed and open. {\sc Bottom:} Nightstand base and communication hardware.} 
\label{fig:nightstand}
\end{figure}

A part of the nightstand was designed to be cylinder with a movable screen covering it (see Figure ~\ref{fig:nightstand}), with circular trays stacked one above the other inside. The trays are rotated using a Wii Nunchuck connected to the system and the door is opened/closed using an IR (infra-red) based proximity sensor (Sharp GP2Y0A21YK). In the design of everyday things \cite{norman2002everyday}, Norman stresses that the \textit{mapping} between intended and actual operations should be natural and visible. We have tried to keep the interactive control as natural and reactive as possible. When the Wii is tilted left, the tray rotates in the clockwise direction (as seen from above) and in the anti-clockwise direction when the Wii is tilted right. Switching between different trays is achieved by selected the `z' and `c' buttons on the Wii. The Wii is also used to control up-down movement of the entire nightstand on a specially constructed base as shown in Figure ~\ref{fig:nightstand}.

\subsubsection{Construction}
The trays (made of cardboard for a prototype) are mounted on sleeves over a central spindle and rotated by continuous servo motors connected to ARDUINOs. A master ARDUINO connects to a laptop that acts as the controller for the robotic nightstand. From the IR and Wii inputs via the ARDUINO (see Figure ~\ref{fig:block_schema}), the laptop knows whether the individual has completed using the nightstand and wants to send it away or not.

\subsection{Detailed Schematic and Communication between Processes} 

\begin{figure}
\center
\begin{tabular}{c}
\placeimgyy{3.5}{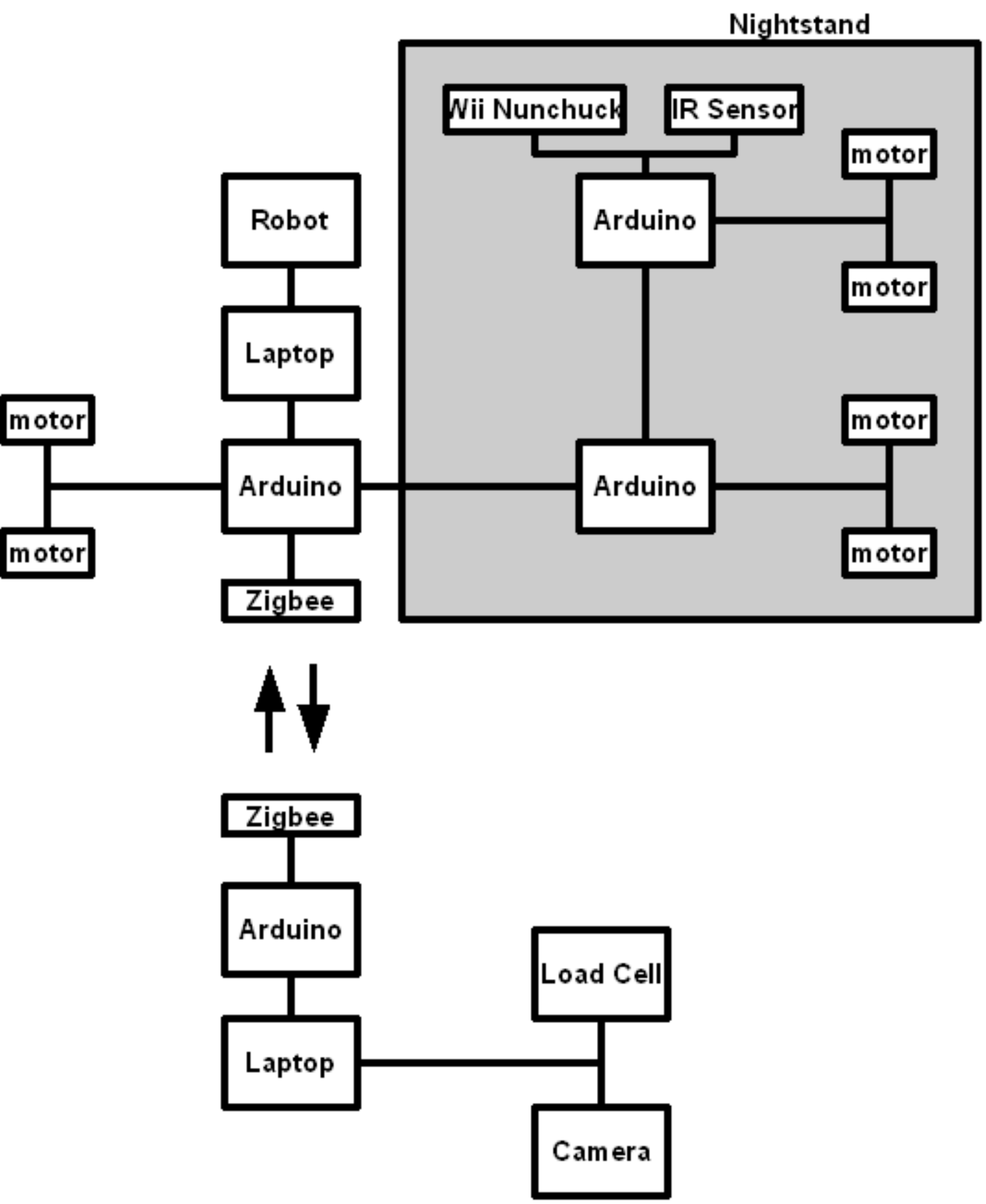}
\end{tabular}
\caption{Detailed Schematic, Perceptual process diagram: The nightstand is mounted on the robot which is controlled by a laptop receiving wireless data through the Xbee (which uses the Zigbee protocol). The mundane controls from the Wii and IR are directly fed in. On the other side, the camera feed and the pressure sensor output are analyzed and required information is transmitted through another Xbee.} 
\label{fig:block_schema}
\end{figure}

\begin{figure}
\center
\begin{tabular}{c}
\placeimgyy{2.0}{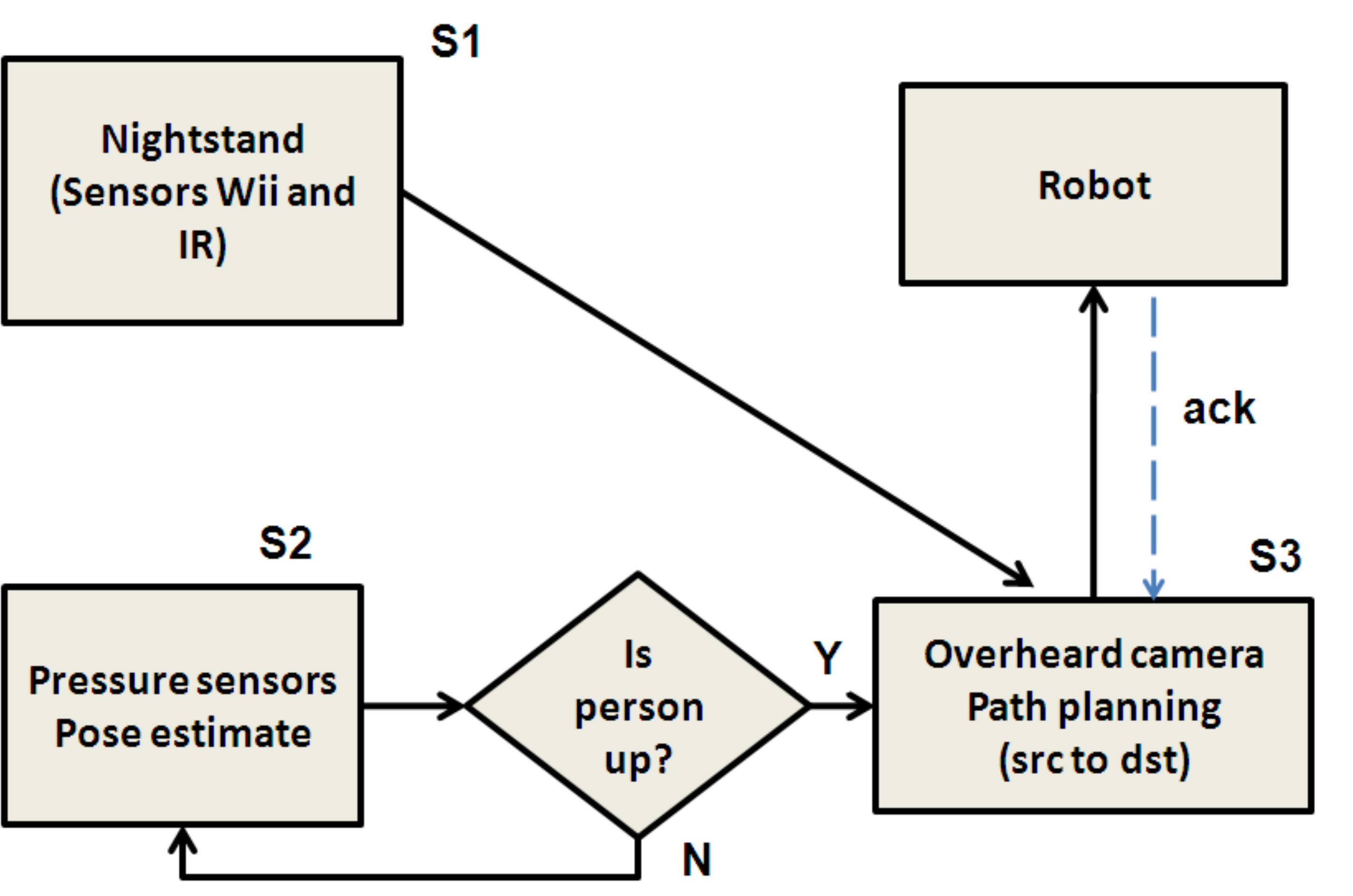}
\end{tabular}
\caption{Communication between the three sensory modules showing integration of perceptual processes.} 
\label{fig:comm}
\end{figure}

See Figure ~\ref{fig:block_schema} for detailed connectivity between the modules. The room intelligence is conveyed to the laptop controlling the robot by means of wireless communication on the ARDUINO. The Xbee shield allows an Arduino board to communicate wirelessly using  the Zigbee protocol. It is based on the Xbee module from MaxStream. The module can communicate up to 100 feet indoors or 300 feet outdoors (with line-of-sight) \cite{arduino}. The communication between different sensory modules is shown in Figure ~\ref{fig:comm}. When the laptop connected to the load cells detects that the person has woken up and is sitting, the camera input is taken and a path is planned for the robot to reach the individual. This information is conveyed to the robot via the wireless Zigbee protocol. The robot approaches the person. When the person has finished using the nightstand, he touches the IR proximity sensor to close the door. This information is conveyed via Zigbee to the laptop obtaining camera feed, which once again plans a path for the robot (with nightstand mounted) to retreat someplace. This information is again conveyed to the robot and the robot retreats. This is just once simple sequence that is possible. Based on similar such sensors deployed in other places in the room, the intelligence in the room could be expanded.

\subsection{Working and Performance}

\subsubsection{HRI Task Metrics addressed during the venture}
A Human-Computer Interaction centric discussion was held in the development team for evaluating and studying the goals of the project which is to be deployed in elderly care homes and medical centers. 

\begin{table}
\begin{center}
\begin{tabular}{|l|c|l|c|l|c|l|c|}
\hline
Task & Addressed\\
\hline
\hline
Navigation: Where it is & Y\\
\hline
Navigation: Where it needs to be & Y\\
\hline
Navigation: Path & Y\\
\hline
Navigation: Obstacles & N\\
\hline
Perception: Search & N\\
\hline
\end{tabular}
\end{center}
\caption{HRI Task metrics}
\label{tab:hri_metrics}
\end{table}

The individual sensory modules have been developed and tested using the ARDUINO platform, Visual C++ and LabVIEW. 
\subsubsection{Pose estimation}
The pose estimation module based on pressure sensors (load cells) resulted in a classification accuracy of 98.5 \% average over 5 predetermined poses (lying, sitting, sitting with one leg up, reclining, sitting up). 15 seconds of data was collected at 10Hz for each individial. Data was collected for 5 poses for 5 individuals. 
\subsection{Nightstand drawers operation}

The use of the Wii for controlling the nightstand operations was tested on 5 graduate students after a short demo. All the students were able to understand and follow the mapping of the motion of the Wii to that of the table drawers. The opening/closing of the door due to proximity was simple too. All the operations worked perfectly in the 100s of trials that were conducted. The Wii and the IR sensors used for mundane operations on the nightstand worked consistently well with a few exceptions in which the gear configurations (using Lego Mindstorms) gave away leading to motor failure.

\subsection{Nightstand robotic base}
The Xbee units were able to successfully communicate bytes with an average delay of 8 ms between each transmission. However we found that this value may vary depending on the ARDUINO module we used. The path planning works crudely well for a simple static environment. Future work would include making the navigation more robust to dynamic changes in the environment. Two sequences were tested for this.
\begin{itemize}
\item Person is lying down and gets up. The pressure sensors detected the motion and sent a signel to the ARDUINO which using XBee initiated the robot to navigate to the person. 
\item When the person laydown again, the pressure sensors again detected and iden tified the action correctly and the robot navigated back to its base. 
\end{itemize}
These sequences were tested atleast 5 times with consistent success. There was one occasion where there was an undue delay in the response of the robot and that was due to a communication delay in the XBee unit that was unexplained. Crude odometry was used for navigation because the drift error was low in the confined space used for testing. In a real environment, odometry would be fused with either visual information or range sensing for more accurate navigation.

\section{Future Work and Discussion}

Future work would involve developing a sophisticated and complete algorithm for path planning and navigation in a dynamic environment. It would be useful to enable a light pattern on the nightstand for it to be detected at night. Also we would like to install a semi-circular railing around the nightstand to assist the individual in walking. The nightstand would also be fitted with an heart-monitor system that checks the pulse-rate of the person when touched. An emergency pill-box would be situated at the base and would be ejected during medical emergencies. 

We have developed an intelligent robotic system with multiple sensors-based perceptual modules. The modules speak to each other to control the robot to navigate to the needs of an aging individual residing at home. We show communication and synergy between the sensors, that work towards monitoring the environment and the person, at all times to enable action without undue user command thereby displaying tropism based on the individuals emergency  needs. There is no need for dialogue, however active interaction is encouraged. The design adheres to the principles of mapping and consistency making the interface natural to the needs of the user. The communication between the three sensory modules shows a natural synergy between the environmental and local sensors. It is also imperative to note that the control is ubiquitous as we would want it to be, as a goal to move one step towards multi-modular, perceptual and pervasive intelligence. 

\bibliography{comforTable}
\bibliographystyle{ieee}
\end{document}